\documentclass{article} % For LaTeX2e
\usepackage{iclr2025_conference,times}

\usepackage{amsmath,amsfonts,bm}

\def\eqref#1{equation~\ref{#1}}
\def\1{\bm{1}}

\DeclareMathAlphabet{\mathsfit}{\encodingdefault}{\sfdefault}{m}{sl}
\SetMathAlphabet{\mathsfit}{bold}{\encodingdefault}{\sfdefault}{bx}{n}

\usepackage{hyperref}
\usepackage{url}
\usepackage{graphicx}
\usepackage{wrapfig}
\usepackage{caption}
\usepackage{subcaption}
\usepackage{booktabs}
\usepackage{diagbox}

\title{MVDrag3D: Drag-based Creative 3D Editing via Multi-view Generation-Reconstruction Priors}

\author{Honghua Chen, Yushi Lan, Yongwei Chen, Yifan Zhou, Xingang Pan %\thanks{ Use footnote for providing further information
\\
S-Lab, Nanyang Technological University\\
}

\iclrfinalcopy % Uncomment for camera-ready version, but NOT for submission.
\begin{document}

\onecolumn{%
\renewcommand\twocolumn[1][]{#1}%
\maketitle
\begin{figure}[h]
    \centering
    \includegraphics[width=\linewidth]{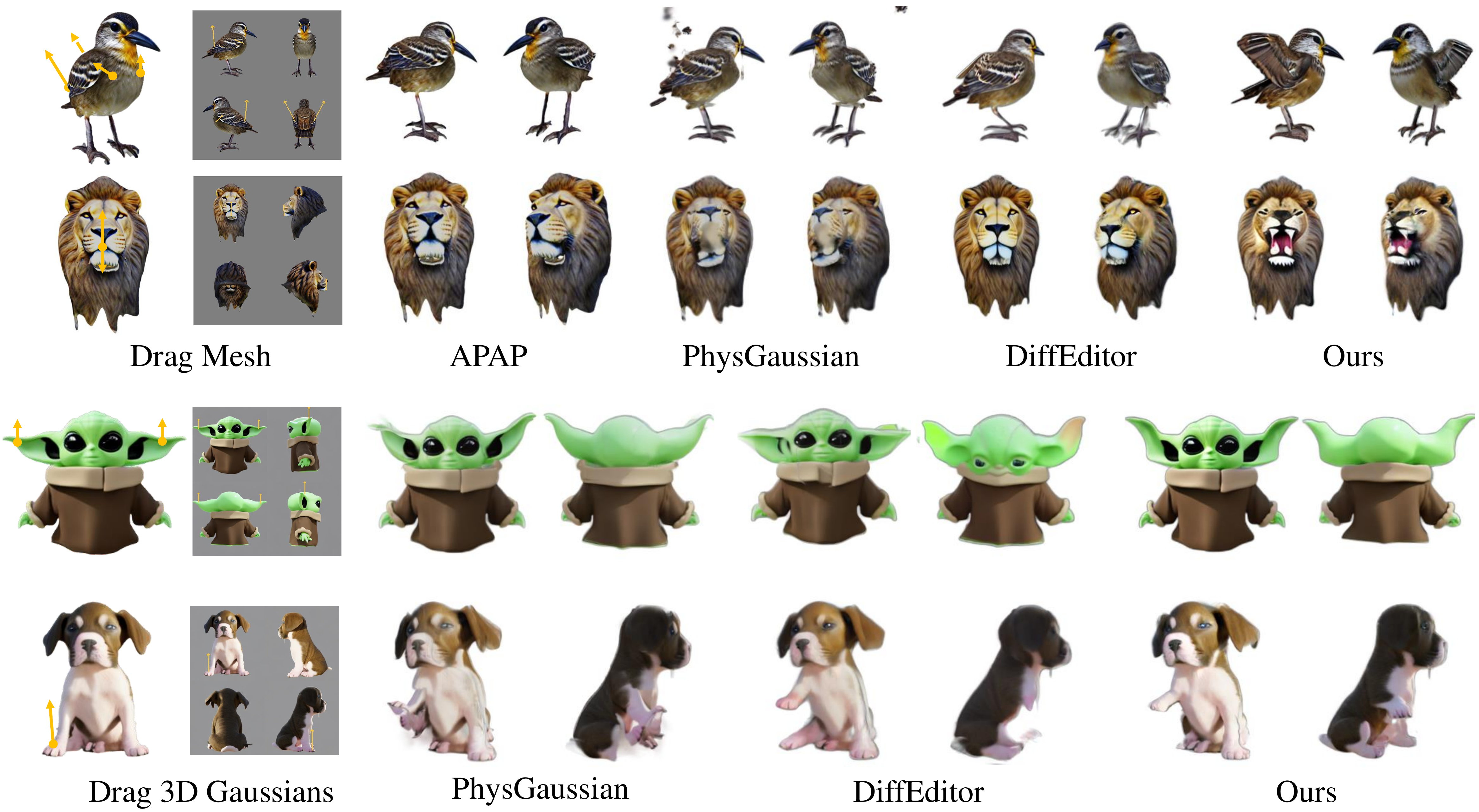}
    \caption{Comparison of our MVDrag3D with state-of-the-art approaches. The first two rows present results of dragging on meshes, while the last two focus on 3D Gaussians. Notably, APAP~\citep{yoo2024plausible} is specifically designed for mesh structures, and thus, it was not tested on 3D Gaussians. Overall, our method demonstrates the ability to produce more plausible and generative editing results, showing better performance across both 3D Gaussians and meshes. }
    \label{fig:teaser}
\end{figure}
}

\begin{abstract}
Drag-based editing has become popular in 2D content creation, driven by the capabilities of image generative models. However, extending this technique to 3D remains a challenge. 
Existing 3D drag-based editing methods, whether employing explicit spatial transformations or relying on implicit latent optimization within limited-capacity 3D generative models, fall short in handling significant topology changes or generating new textures across diverse object categories.
To overcome these limitations, we introduce \textit{\textbf{MVDrag3D}}, a novel framework for more flexible and creative drag-based 3D editing that leverages multi-view generation and reconstruction priors.
At the core of our approach is the usage of a multi-view diffusion model as a strong generative prior to perform consistent drag editing over multiple rendered views, which is followed by a reconstruction model that reconstructs 3D Gaussians of the edited object.
While the initial 3D Gaussians may suffer from misalignment between different views, we address this via view-specific deformation networks that adjust the position of Gaussians to be well aligned.
In addition, we propose a multi-view score function that distills generative priors from multiple views to further enhance the view consistency and visual quality. Extensive experiments demonstrate that MVDrag3D provides a precise, generative, and flexible solution for 3D drag-based editing, supporting more versatile editing effects across various object categories and 3D representations. Video demos can be found on our project webpage: \href{https://chenhonghua.github.io/MyProjects/MvDrag3D/}{\textit{https://chenhonghua.github.io/MyProjects/MvDrag3D/}}.

\end{abstract}
\section{Introduction}
% 1. Drag-based editing scheme is very important in 3D
%Drag-based 3D editing has long been an essential interactive tool in computer graphics, enabling intuitive manipulation of complex shapes and structures through simple point-based interactions. 
Deforming 3D shapes by dragging point handles has been an essential interactive tool in computer graphics, enabling intuitive manipulation of complex shapes and structures.
Traditionally, such drag-based 3D editing is often defined on mesh structures, utilizing optimization functions to preserve specific properties under the constraint of control handles. These properties include the mesh Laplacian~\citep{lipman2004differential, lipman2005linear, sorkine2004laplacian}, local rigidity~\citep{igarashi2005rigid, sorkine2007rigid}, and surface Jacobians~\citep{aigerman2022neural, gao2023textdeformer}, as well as more recent considerations of perceptual plausibility~\citep{yoo2024plausible}. 
However, these methods are constrained by the fixed topology of mesh structures, limiting their flexibility, especially in complex edits that require substantial changes to the topology or the generation of new textures, e.g., editing a bird to open its wings. %(see the results of APAP~\citep{yoo2024plausible} in Fig.~\ref{fig:teaser}).

%Since 3D Gaussian splatting~\citep{kerbl20233d} is more expressive and retains an explicit structure that is easy for user interaction, 
In light of the recently introduced 3D Gaussian splatting~\citep{kerbl20233d} that is more expressive and easy to edit, 
Interactive3D~\citep{dong2024interactive3d} introduces a series of deformable and rigid 3D operations to directly manipulate local 3D Gaussians. This is followed by Gaussian-to-NeRF reformatting and refinement through Score Distillation Sampling (SDS)~\citep{poole2022dreamfusion}. However, this method suffers from prolonged NeRF optimization and the typical limitations of vanilla SDS, such as over-saturation.
PhysGaussian~\citep{xie2024physgaussian} also simulates drag-induced motion by integrating physically grounded dynamics into 3D Gaussians. However, it requires an accurate predefinition of the physical properties involved, which can be difficult to obtain. 
Besides, both methods still face challenges in making large structural changes and generating new content.
%refining details or generating coherent new content.

Notably, recent drag-based editing has seen considerable success in the 2D domain~\citep{pan2023drag,mou2023dragondiffusion,mou2024diffeditor,zhang2024gooddrag,shin2024instantdrag}, largely due to the capabilities of powerful image generative models, such as GANs~\citep{karras2020analyzing} and diffusion models~\citep{rombach2022high}. These models encompass a latent space that enables various harmonious manipulations, including object deformation, layout adjustments, and coherent new content generation.
Building on this success, some 3D editing methods have begun to explore generative 3D dragging within a 3D latent space. For instance, Drag3D~\citep{ashawkeyDrag3D2023}, adapts DragGAN~\citep{pan2023drag} by incorporating a 3D GAN~\citep{shen2021deep} into a motion-based latent optimization framework. %However, its performance is constrained by the capacity and generalization of current 3D GANs. 
Similarly, CNS-Edit~\citep{hu2024cns} employs a latent-based method but combines it with a 3D neural volume diffusion model~\citep{hui2022neural}. This approach requires training separate models for each shape category, making it less flexible and more resource-intensive.
Obviously, both of the above approaches are limited by the capacity and generalization of current 3D generative models.

%Interestingly, 
In pursuit of a stronger generative prior for more powerful drag-based 3D editing, we have observed the following from existing 3D generation and reconstruction work:
1) most 3D representations can be rendered into multiple views;
2) 3D objects can be faithfully reconstructed from four and more views~\citep{tang2024lgm,xu2024grm}; and
3) existing multi-view diffusion models provide a strong prior for generating consistent images across four orthogonal views~\citep{shi2023mvdream,kant2024spad}.
These observations inspire us to explore the potential of leveraging both \textit{large-scale multi-view generation and reconstruction models} as 3D priors, agnostic to 3D representations, to facilitate precise, generative, and general 3D dragging.
Ideally, we expect that the 3D dragging operation should exhibit the following properties 
1) \textit{Accuracy}: the ability to precisely drag any point on a 3D object's surface to a target spatial position;
2) \textit{Generative capability}: the ability to generate visually plausible new content to match the drag intention; and
3) \textit{Versatility}: compatibility with various input object categories and most 3D representations, such as 3D Gaussians or meshes.

% 4. what we do
To this end, we introduce MVDrag3D, a novel framework for drag-based 3D editing that leverages multi-view generation and reconstruction priors. Our method begins by rendering four orthogonal views of a 3D object and projecting the dragging points onto the corresponding views. To ensure consistent 3D edits, we extend the score-based gradient guidance mechanism within a multi-view diffusion model and propose a multi-view guidance energy function, enabling consistent edits across all four views.
Thanks to the generative capabilities of the multi-view diffusion model, edits across four views can faithfully reflect significant structural changes or newly synthesized textures.
The edited views are then fused into a 3D Gaussian representation using a multi-view Gaussian reconstruction model. Although the initial 3D Gaussian appears complete, we observe a loss of appearance detail, and the 3D Gaussians in the overlapping regions between views do not align accurately, leading to noticeable discrepancies in the 2D rendering. To address these issues, we employ a deformation network that predicts the displacement of each Gaussian to correct the 3D alignment. Additionally, we formulate an image-conditioned multi-view score function to distill generative priors from the multiple views simultaneously, ensuring high-fidelity results while preserving details across all views.
We summarize our contributions as follows:

\begin{enumerate}
    \item We propose MVDrag3D, a drag-based 3D editing framework that leverages multi-view generation-reconstruction priors. It is accurate, generative, and adaptable to diverse input categories and most 3D representations, such as 3D Gaussians and meshes.
    \item We extend the gradient guidance mechanism into a multi-view diffusion model and introduce multi-view guidance energy, which ensures consistent drag-based edits across four views.
    \item We design a lightweight deformation network that corrects each 3D Gaussian's position and enhances geometric consistency. Furthermore, we introduce an image-conditioned multi-view score function to iteratively refine the 3D Gaussian, ensuring high-fidelity appearance and preserving fine details across all views.
\end{enumerate} 

\section{Related work}
We will review prior research, starting from drag-based 2D image editing techniques, and progressing to more recent developments in drag-based 3D editing and 3D generation-reconstruction priors.

% \subsection{Drag-based image editing}
\textbf{Drag-based image editing}.
Drag-based image manipulation allows users to exert precise control over specific areas of the image via manual interactions like dragging and clicking. Most existing techniques employ iterative latent optimization in the latent space, and they can be roughly divided into two categories: methods that rely on motion tracking~\citep{pan2023drag, shi2023dragdiffusion, zhang2024gooddrag, Cui2024StableDragSD, liu2024drag, ling2023freedrag} and those based on guidance gradients~\citep{mou2023dragondiffusion, mou2024diffeditor}.
DragGAN~\citep{pan2023drag}, for instance, optimizes the latent space of GANs using iterative motion supervision and point tracking. Later, diffusion-based methods, including DragDiffusion~\citep{shi2023dragdiffusion}, GoodDrag~\citep{zhang2024gooddrag}, StableDrag~\citep{Cui2024StableDragSD}, DragNoise~\citep{liu2024drag}, and FreeDrag~\citep{ling2023freedrag}, have further refined these motion-driven techniques for more refined results.
Meanwhile, DragonDiffusion~\citep{mou2023dragondiffusion} and DiffEditor~\citep{mou2024diffeditor} utilize a gradient-based approach by optimizing an energy function~\citep{epstein2023diffusion} to achieve desired edits. Since both motion- and gradient-based methods require time-consuming iterations, SDEDrag~\citep{nie2024the} and FastDrag~\citep{zhao2024fastdrag} have been proposed to accelerate the editing process. More recently, InstantDrag~\citep{shin2024instantdrag} decomposes the dragging task into two components: learning motion dynamics and generating images conditioned on motion, achieving a better balance among interactivity, speed, and quality. %For drag-based 3D editing, since a 3D object can be rendered into multiple images, we can edit each view independently and then reconstruct the 3D model. However, this often leads to significant 3D inconsistencies. In this paper, building on the insights from previous work, particularly the gradient-guided editing technique, we extend it into the 3D domain to achieve both precise control and generative flexibility, while maintaining 3D consistency.

% \subsection{Drag-based 3D editing}
\textbf{Drag-based 3D editing}.
To achieve drag-based 3D editing, classical mesh deformation techniques are commonly employed. These methods often design optimization functions to preserve specific geometric properties, such as the mesh Laplacian~\citep{lipman2004differential, lipman2005linear, sorkine2004laplacian}, local rigidity~\citep{igarashi2005rigid, sorkine2007rigid}, and surface Jacobians~\citep{aigerman2022neural, gao2023textdeformer}, under the constraints of user-interactive handles like key points or cages. Despite their widespread use, these techniques frequently result in unnatural shape distortion, primarily due to their inability to ensure perceptual plausibility. To address this limitation, APAP~\citep{yoo2024plausible} introduced an innovative approach by incorporating SDS loss to optimize the Jacobian deformation field. However, like previous mesh deformation methods, APAP is constrained by the fixed topology of mesh structures, limiting its flexibility, particularly for complex edits that require generating entirely new content. On the other hand, Interactive3D~\citep{dong2024interactive3d} introduces a series of deformable and rigid 3D point operations on 3D Gaussians and also employs SDS to optimize the deformed or transformed Gaussians/NeRFs. Besides, PhysGaussian~\citep{xie2024physgaussian} also involves certain types of drag-related motion by integrating physically grounded dynamics into 3D Gaussians, however, it requires a suitable predefinition of the physics involved. 
Although these latter two methods employ more expressive 3D representations, they often require labor-intensive post-processing and face challenges in refining fine details or generating coherent new content.

As drag-based image editing techniques evolve, some 3D editing methods have begun to explore generative 3D dragging within a 3D latent space. For instance, Drag3D~\citep{ashawkeyDrag3D2023}, built upon DragGAN~\citep{pan2023drag}, integrates a 3D GAN model into a motion-based latent optimization framework. However, the approach is inherently limited by the capacity and generalization constraints of current 3D GAN models. Later, CNS-Edit~\citep{hu2024cns} introduces a coupled neural shape representation to facilitate 3D shape editing. This method utilizes a latent code to capture high-level global semantics, while a 3D neural feature volume provides spatial context for local shape modifications. However, CNS-Edit's category-specific design requires separate models for different 3D shape categories.  Different from them, in this work, we achieve 3D generative dragging within a more powerful multi-view latent space.

% \subsection{Multi-view generation-reconstruction priors}
% \textbf{Multi-view generation-reconstruction priors}.
\noindent\textbf{Multi-view Image Generation}.
2D diffusion models~\citep{rombach2022high,saharia2022photorealistic} initially focus on generating a single-view image. 
Recently, several models~\citep{shi2023mvdream,wang2023imagedream,shi2023zero123++,li2023sweetdreamer,long2024wonder3d,kant2024spad,tang2024mvdiffusion++,liu2024one} turned to employ a 3D-aware multi-view diffusion approach, incorporating camera poses as additional inputs and fine-tuning the diffusion model on multi-view data~\citep{deitke2023objaverse}. This strategy enables the consistent generation of multi-view images representing the same object.
Essentially, these multi-view diffusion models capture a rich, generalizable distribution of 3D data, agnostic to a specific 3D representation. Also, given the limitations of current ``pure'' 3D generative models—those trained directly on 3D data—we believe that leveraging multi-view diffusion models as a 3D prior proxy could offer a promising solution for flexible 3D editing. 

\noindent\textbf{Feed-forward Multi-view 3D Reconstruction}.
By generating 3D-consistent multi-view images, various optimization techniques can be employed to reconstruct 3D objects~\citep{shi2023mvdream,wang2023imagedream,liu2023syncdreamer}. To improve generation speed and quality, more recent work has explored large-scale reconstruction models using multi-view images (e.g., 4 or 6)~\citep{wang2023pf,xu2023dmv3d,li2023instant3d,wang2024crm,xu2024instantmesh}. These approaches leverage transformers to directly regress triplane-based NeRF representations. Newer methods like LGM~\citep{tang2024lgm} and GRM~\citep{xu2024grm} replaced triplane NeRF with 3D Gaussians~\citep{kerbl20233d}, achieving high-fidelity rendering at faster speeds.
In summary, these recent feed-forward multi-view reconstruction models provide a robust 3D reconstruction prior, enabling the fast and faithful recreation of complete 3D objects from sparse-view images. In this work, we utilized a 4-view reconstruction model~\citep{tang2024lgm} and a 4-view diffusion model~\citep{shi2023mvdream} as our generation-reconstruction priors.
%, and an image-conditioned 4-view diffusion model~\citep{wang2023imagedream}

\section{Method}
In this section, we briefly introduce score-based guidance energy for image editing, followed by a detailed explanation of our method.

\subsection{Preliminary}
\textbf{Score-based gradient guidance for image editing.} 
Recently, DragonDiffusion~\citep{mou2023dragondiffusion} and DiffEditor~\citep{mou2024diffeditor} have applied score-based gradient guidance~\citep{dhariwal2021diffusion} to efficient and flexible image-editing tasks. The score function enables sampling from a more enriched distribution, generally defined as: 
\begin{equation}
\label{eq:self_guidance_score}
\tilde{\bm{\epsilon}}_{\theta}^t(\mathbf{x}_t)=\bm{\epsilon}_{\theta}^t(\mathbf{x}_t) + \eta\cdot\nabla_{\mathbf{x}_t} \mathcal{E}(\mathbf{x}_t, \mathbf{y}),
\end{equation}
where the first term is the unconditional denoiser, and the second term is the conditional gradient produced by an energy function. Here, $\eta$ is the learning rate, and $\mathbf{y}$ represents the edit target, such as text embedding. During the diffusion sampling process, the gradient guidance from the energy function aligns with the editing target, gradually modifying the input image to meet the desired edit.

In recent 2D dragging task~\citep{mou2024diffeditor,mou2023dragondiffusion}, the guidance energy function is constructed based on image feature correspondence within a pre-trained diffusion model as follows:
\begin{equation}
\small
\label{eq:self_guidance_energy}
    \nabla_{\mathbf{z}_t} \log q(\mathbf{y}|\mathbf{z}_t) = \alpha \cdot \mathbf{m}_{edit}\cdot \nabla_{\mathbf{x}_t}\mathcal{E}_{edit} + \beta \cdot (1-\mathbf{m}_{edit})\cdot \nabla_{\mathbf{x}_t}\mathcal{E}_{content},
\end{equation}
where $\mathbf{m}_{edit}$ is the editing region mask. The energy function $\mathcal{E}_{edit}$ measures the diffusion feature similarity between areas near the dragging start and destination points, while $\mathcal{E}_{content}$ ensures that unedited content stays consistent with the original image. $\alpha$ and $\beta$ are balance weights.
In our work, we extend both the editing energy and content energy to a multi-view version. This ensures that modifications made in one view are coherently reflected across all views.

\begin{figure*}[t]
    \centering
    \includegraphics[width=\linewidth]{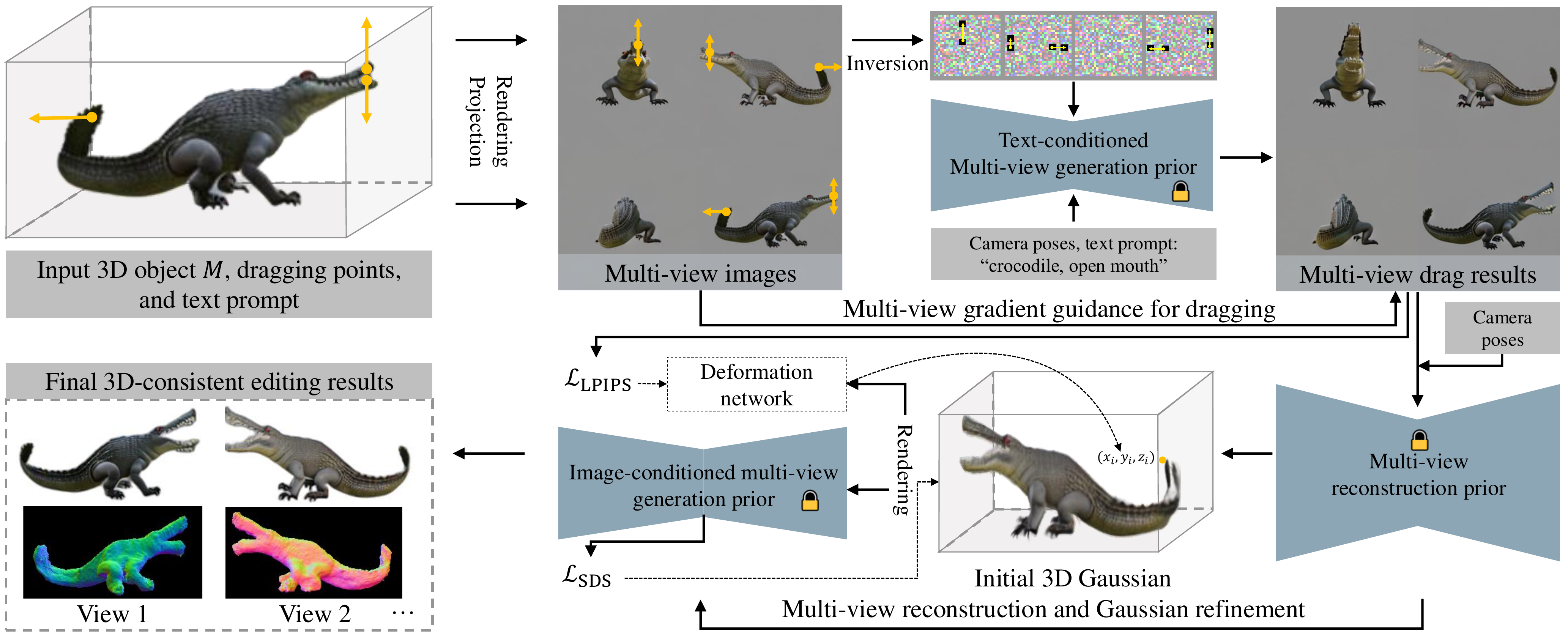}
    \caption{Method overview. Given a 3D model and multiple pairs of 3D dragging points, we first render the model into four orthogonal views, each with corresponding projected dragging points. Then, to ensure consistent dragging across these views, we define a multi-view guidance energy within a multi-view diffusion model. The resulting dragged images are used to regress an initial set of 3D Gaussians. Our method further employs a two-stage optimization process: first, a deformation network adjusts the positions of the Gaussians for improved geometric alignment, followed by image-conditioned multi-view score distillation to enhance the visual quality of the final output.}
    \label{fig:overview}
\end{figure*}

\subsection{Overview}
\label{sec:problem}
The entire process is visualized in Fig.~\ref{fig:overview}. Given a 3D model $M$ to be edited, and $k$ pairs of 3D dragging points $\{(\mathbf{p}_j^{3D}, \mathbf{q}_j^{3D})\}_{j=1}^{k}$, we first render $M$ into four orthogonal images $\mathcal{I} = \{\mathbf{I}_i\}_{i=1}^{4}$, along with the corresponding dragging points (Sec.~\ref{sec:2D_projection}). 
We then propose a multi-view guidance energy function (Sec.~\ref{sec:mvdrag}), which ensures consistent and coherent dragging across all views. The edited images $\mathcal{I}_e = \{\mathbf{I}_{e,i}\}_{i=1}^{4}$ are used to regress 3D Gaussians using~\citep{tang2024lgm}. While the initial reconstruction appears complete, we further use a deformation network and introduce an image-conditioned multi-view score distillation to correct the misalignment between Gaussians in the overlapping regions of each view and enhance the visual appearance across all views, resulting in the final edited results (represented in 3D Gaussians) (Sec.~\ref{sec:reconstruction}).

\subsection{3D-2D Rendering and Projection}
\label{sec:2D_projection}
We decompose the 3D dragging operation in a multi-view manner. First, we render the 3D model $M$ into four orthogonal images $\{\mathbf{I}_i\}_{i=1}^{4}$ using any suitable renderer. Since MVDream typically generates images with gray backgrounds, we adopt a similar gray background for rendering. 
In terms of camera setup, we adopt the same configuration as MVDream~\citep{shi2023mvdream} and LGM~\citep{tang2024lgm}, which serve as our generation-reconstruction priors. Specifically, the four views are chosen at orthogonal azimuths $(0^{\circ}, 90^{\circ}, 180^{\circ}, 270^{\circ})$ and a fixed elevation $(0^{\circ})$. Then, the $k$ pairs of 3D dragging points can be projected onto the corresponding views, represented as $\{(\mathbf{p}_{i,j}^{2D}, \mathbf{q}_{i,j}^{2D})\}_{j=1}^{k}$. However, due to potential occlusions in certain views, we discard the point pairs if the $z$-axis value of $\mathbf{p}_{i,j}^{2D}$ or $\mathbf{q}_{i,j}^{2D}$ exceeds the rendered depth at the corresponding 2D position.

\subsection{Multi-view gradient guidance for dragging}
\label{sec:mvdrag}
\begin{figure*}[t]
    \vspace{-2mm}
    \centering
    \includegraphics[width=\linewidth]{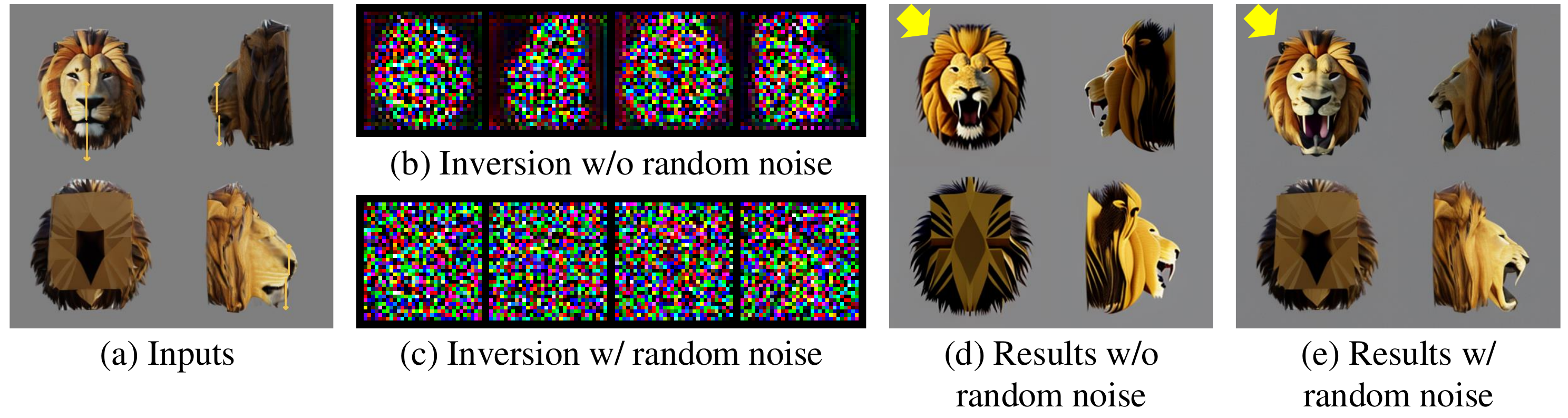}
    \caption{Effect of DDIM inversion with random noise. For the rendered four images, when inverted into MVDream's data distribution, the resulting noise deviates from a Gaussian distribution (b). By adding random noise ($\mathcal{N}(0, 0.01)$) to the background's pixel domain, we help the latent variables conform more closely to a Gaussian distribution (c). The resulting multi-view edits are shown in (d) and (e). Yellow arrows indicate the views with evident identity changes.}
    \label{fig:ddim}
    \vspace{-4mm}
\end{figure*}

Since a 3D object can be rendered into multiple images and numerous drag-based 2D editing methods already exist, a straightforward approach to achieve drag-based 3D editing would be to independently edit each view and then reconstruct the 3D model. However, this leads to significant 3D inconsistencies (see the results of DiffEditor~\citep{mou2024diffeditor} in Fig.~\ref{fig:teaser}), as the editing results of each image become misaligned across various factors such as pose, layout, texture, and more.
Based on the observation that multi-view diffusion models can simultaneously generate a consistent set of multi-view images, and recognizing the effectiveness of score-based gradient guidance in image editing, we extend gradient guidance to a multi-view version. 

Specifically, we first apply DDIM inversion~\citep{song2020denoising} to transform each of $\{\mathbf{I}_i\}_{i=1}^{4}$ into a Gaussian distribution. These distributions are combined and represented as $\mathbf{z}_{T} \in \mathcal{R}^{4 \times H \times W \times C}$ within the latent space of MVDream. Using $\mathbf{z}_{T}$, we can extract an intermediate feature $\mathbf{F}$ from the UNet decoder.
Note that MVDream reshapes $\mathbf{z}_{T}$ into a $4HW \times C$ format, thus extending self-attention to the cross-view version. This ensures that guidance from one view can influence the others. 
With this, we follow~\citep{mou2023dragondiffusion} and define a multi-view guidance energy:
\begin{equation}
\label{eq:mv_edit_energy}
\begin{aligned}
    \mathcal{E}_{edit} = \sum_{i=1}^{4}\frac{1}{0.5\cdot \cos\left(\mathbf{F}_{i, t}^{edi}[\mathbf{m}^{edi}_i],\ sg(\mathbf{F}_{i, t}^{ori}[\mathbf{m}^{ori}_i])\right)+0.5}, \\
    \mathcal{E}_{content} = \sum_{i=1}^{4}\frac{1}{0.5\cdot \cos\left(\mathbf{F}_{i, t}^{edi}[\mathbf{m}^{unedited}_i],\ sg(\mathbf{F}_{i, t}^{ori}[\mathbf{m}^{unedited}_i])\right)+0.5},
\end{aligned}
\end{equation}
where $\mathbf{F}_{i, t}^{edi}$ and $\mathbf{F}_{i, t}^{ori}$ are intermediate features of $\mathbf{z}_{i, t}^{edi}$ and $\mathbf{z}_{i, t}^{ori}$. $\mathbf{z}_{i, t}^{ori}$ corresponds to the latent variables of original image at time step $t$, while $\mathbf{z}_{i, t}^{edi}$ represents the edited latent variable. $sg(\cdot)$ is the gradient clipping operation. 
In the dragging operation, 
%$\mathbf{z}_{t}^{edi}$ is initially set equal to $\mathbf{z}_{t}^{ori}$, except within the dragging destination region, where latents from the starting region are copied to replace the original values. 
$\mathbf{m}^{ori}$ (or $\mathbf{m}^{edi}$) is a $3 \times 3$ rectangular patch centered around the 2D dragging points $\mathbf{p}^{2D}$ (or $\mathbf{q}^{2D}$). $\mathbf{m}^{unedited}$ denotes the areas without editing.
To enhance readability, the index labels on each image are omitted.
Note also that all layers of the UNet decoder features are used to compute the guidance energy, ensuring more comprehensive and robust results. The gradient of $\mathcal{E}_{edit}$ is then used to generate consistently edited images $\{\mathbf{I}_{e,i}\}_{i=1}^{4}$, while $\mathcal{E}_{content}$ employed to preserve the appearance of the unedited regions, keeping them as close to the original images as possible.

\textbf{DDIM inversion with random noise}. During DDIM inversion, we observed that for the given four images, their latent noise does not follow a Gaussian distribution, as depicted in Fig.~\ref{fig:ddim} (b). 
This discrepancy often causes instability during the editing process, making it difficult to preserve the object's identity (see Fig.~\ref{fig:ddim} (d)). We believe this issue arises because MVDream was never trained on images with smooth, noise-free regions like the background, leading to a domain gap during inversion~\citep{ouyang2024i2vedit}.
To address this issue, we found that introducing small, nearly imperceptible perturbations to the pixel domain—especially in smooth areas like the background—significantly improves the inversion process. These subtle disturbances help the latent variables conform more closely to a Gaussian distribution (see Fig.~\ref{fig:ddim} (c)). The final results exhibit smoother transitions and better overall fidelity in the edited images, as shown in Fig.~\ref{fig:ddim} (e).

\subsection{3D Gaussian Reconstruction and Refinement}
\label{sec:reconstruction}

\begin{wrapfigure}{r}{0.5\textwidth}
    \vspace{-6mm}
    \begin{center}
       \includegraphics[width=\linewidth]{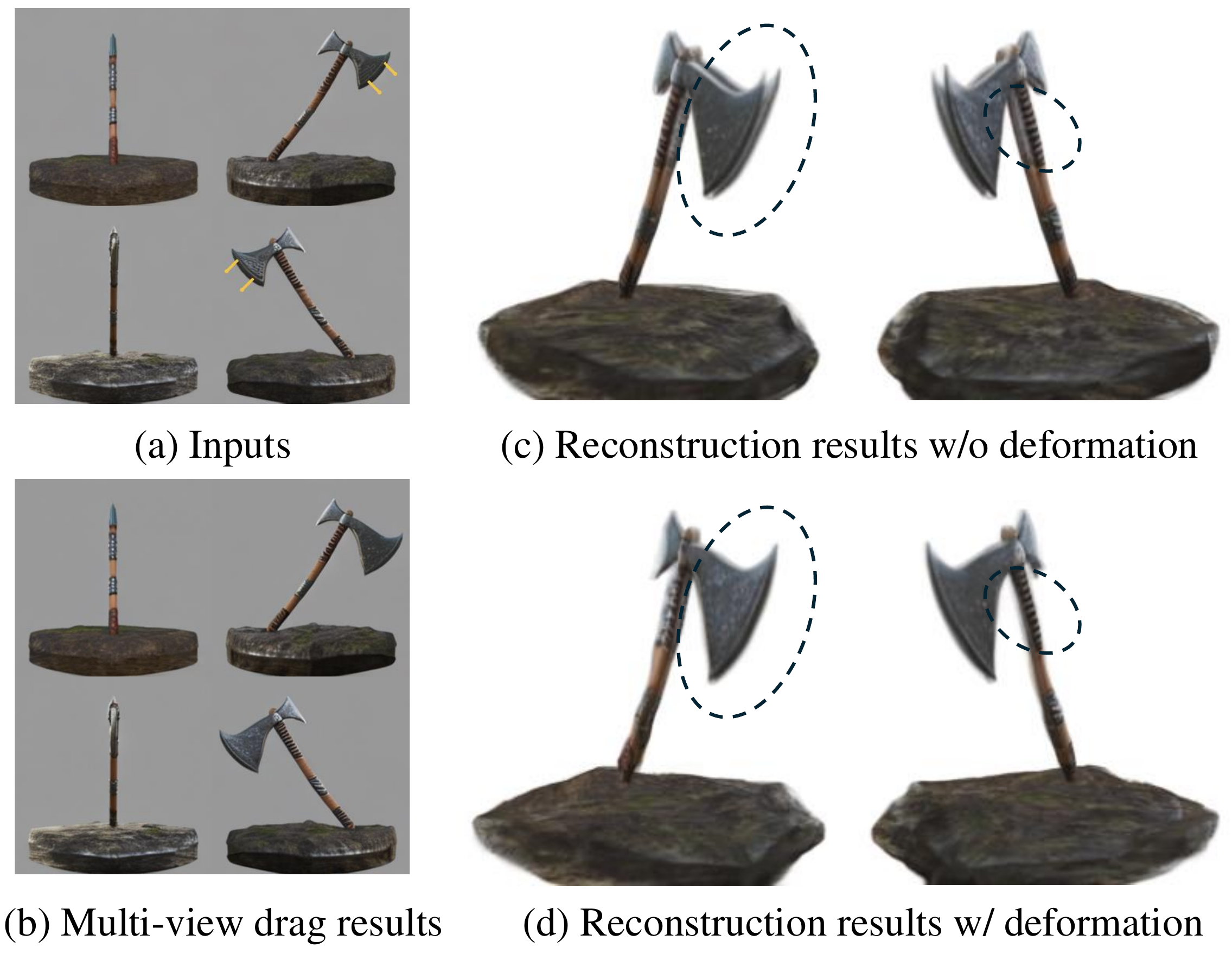}
        \caption{Effect of Gaussian position optimization. (c) shows 3D reconstruction result may exhibit structural misalignment. By employing a deformation network to optimize the Gaussian position, we achieve better compactness and consistency among the Gaussians across different views, as shown in (d).
        }
        \label{fig:deformation}
    \end{center}
    % \vspace{-6mm}
\end{wrapfigure}

\begin{wrapfigure}{r}{0.5\textwidth}
    \vspace{-4mm}
    \begin{center}
       \includegraphics[width=\linewidth]{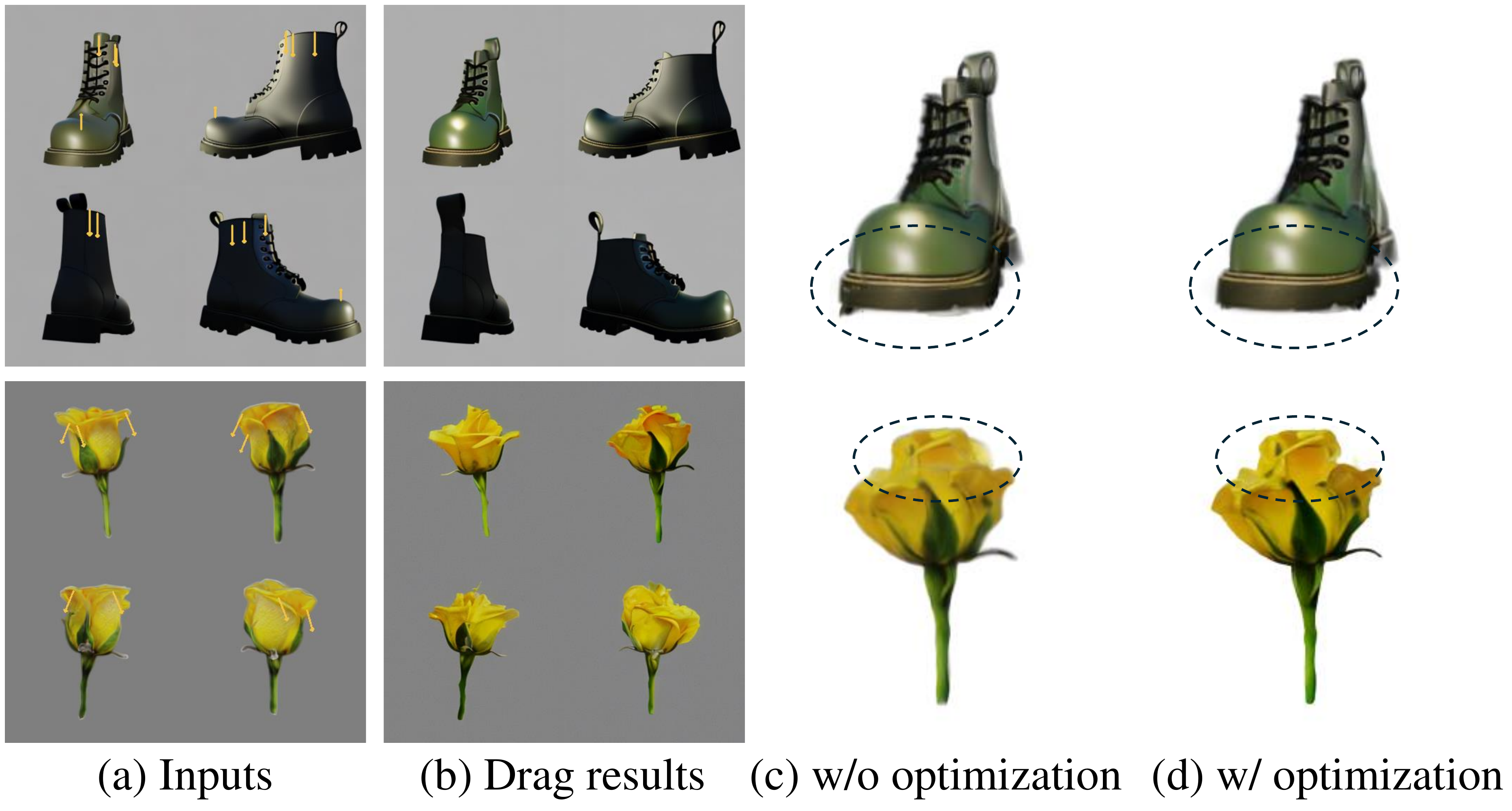}
        \caption{Effect of image-conditioned multi-view SDS. (c) presents the reconstruction results without appearance optimization, while (d) displays the corresponding results after optimization, which are noticeably sharper and clearer.}
        \label{fig:sds}
    \end{center}
    % \vspace{-4mm}
\end{wrapfigure}

Once we obtain the four edited images, we employ LGM~\citep{tang2024lgm} to regress a partial 3D Gaussians for each view and then fuse them into a unified 3D Gaussian representation. However, we encountered two significant challenges: (1) because we only use four orthogonal views, the predicted Gaussians in the overlapping regions between views are usually not aligned correctly, resulting in noticeable discrepancies in the 2D rendering (see Fig.~\ref{fig:deformation} (c)), and (2) the appearance details are frequently lost during LGM's regression process, reducing the visual fidelity of the final 3D reconstruction (see Fig.~\ref{fig:sds} (c)).

In our early tests, to address these issues, we applied vanilla SDS on the initial reconstruction, incorporating a multi-view reconstruction loss across the four views. However, these adjustments did not resolve the underlying issues. 
We attribute these challenges to the inherent ambiguity in the SDS and reconstruction losses. Specifically, it is difficult to directly optimize independent Gaussians consistently without regularization, and the losses do not effectively indicate when to adjust the position or when to densify or prune the Gaussians, resulting in suboptimal outcomes.
To address these challenges, we propose a two-step approach: first, we adjust the Gaussian's position via deformation fields to achieve better geometric alignment and then focus on enhancing visual quality.

\textbf{Gaussian position optimization}. 
Considering that the geometric misalignment problem across views mainly involves low-frequency overall structural changes and the Gaussians belonging to the same view should be moved more consistently, for each view' Gaussian set, we propose to use an individual deformation network $f$ to predict each Gaussian's movement $(\delta x_i, \delta y_i, \delta z_i)$. This means we employ a total of four lightweight individual MLPs, one for each view. Besides, since standard MLPs are generally ineffective for low-dimensional coordinate-based regression tasks~\citep{tancik2020fourier}, we enhance the model by applying Fourier positional embeddings ($pe(\cdot)$) to each Gaussian's $(x, y, z)$ coordinates. The new position for each Gaussian is then calculated as: $(x^\prime, y^\prime, z^\prime) = (x, y, z) + f(pe((x, y, z)))$. 
The training loss is the VGG-based LPIPS loss, applied to the four images. This helps maintain perceptual similarity and ensures better alignment across views: $\mathcal L_\text{LPIPS} = \sum_{i=1}^{4} \text{LPIPS} (\mathbf{I}_{e,i}, \mathbf{I}^\text{render}_{e,i}), \label{eq:deformation_loss}$  
where $\mathbf{I}^\text{render}_{e,i}$ is the rendered image by the optimized Gaussians after their positions have been corrected. Note that Gaussian densification and pruning are not performed at this stage. Fig.~\ref{fig:deformation} (d) shows the effectiveness of the Gaussian position optimization stage.

\textbf{Gaussian appearance optimization}. 
The deformation network described above is limited to optimizing the positions of the Gaussians. When extending MLPs to optimize other Gaussian properties, such as spherical harmonics, we observe no significant improvement in appearance details. Inspired by ReconFusion~\citep{wu2024reconfusion}, we propose to frame the Gaussian appearance enhancement task as an image-conditioned multi-view SDS optimization problem.
Our objectives are two-fold: (1) ensuring multi-view consistency across novel camera angles beyond the initial four views and (2) preserving the identity of the edited four views. To achieve this, we define the edited-image conditioned multi-view score function:

\begin{equation}
    \nabla_{\phi} \mathcal{L}_\textrm{SDS} =  \mathbb{E}_{t, \epsilon, o} [(\epsilon_\theta(\hat{I}; t, \mathbf{I}_{e,i},o) - \epsilon)\frac{\partial \hat{I}}{\partial \phi}], \text{and } i=1, 2, 3, \text{or }4,
    \label{eq:sds}
\end{equation}

where $\hat{I}$ represents the rendered batch images from any four orthogonal views, and $o$ denotes the corresponding camera poses. During each SDS iteration, we randomly render four orthogonal views and randomly select one edited image $\mathbf{I}_{e,i}$ as a condition to compute the SDS loss. The multi-view diffusion model employed is ImageDream~\citep{wang2023imagedream}, which can be seen as an image-conditioned version of MVDream. This allows it to be seamlessly integrated into our framework. 
In each iteration, we also compute $\mathcal{L}_\text{LPIPS}$. Note that all Gaussian properties are optimized during this process, with densification and pruning operations enabled.

\section{Experiments}
\subsection{Experimental Setup}
\textbf{Implementation Details}. We conducted all experiments on a single 48 GB A6000 GPU. For multi-view image dragging, we employed DDIM sampling with 150 steps, applying random Gaussian noise $\mathcal{N}(0,0.01)$ to the background. 
In the Gaussian deformation stage, we used $4$ MLPs, each trained for $2,000$ iterations with a learning rate of $0.00001$. Each MLP consists of a linear layer, a ReLU activation, and another linear layer arranged in a residual structure. For multi-view SDS optimization, we performed $1,000$ iterations, gradually decaying $T_\mathrm{max}$ from $0.49$ to $0.02$.

\textbf{Datasets}. We perform dragging on two of the most popular 3D representations: meshes and 3D Gaussians. For the mesh experiments, we collected $8$ meshes from~\citep{yoo2024plausible} and \textit{Genie}~\citep{lumi}. For the 3D Gaussian experiments, we collected $8$ 3D Gaussians from~\cite{tang2024lgm}. 
We collect data that are representative to demonstrate drag editing but do not cherry-pick based on any results.
The 3D drag points are manually specified using MeshLab, following~\citep{yoo2024plausible}.

\textbf{Metrics}. In this work, we employ two assessment metrics for quantitative evaluation: Dragging Accuracy Index (\textbf{DAI})~\citep{zhang2024gooddrag} and \textbf{GPTEval3D}~\citep{wu2024gpt}. DAI measures the effectiveness of a method in transferring source content to a target point. While DAI effectively measures drag accuracy, it is insufficient because the editing process sometimes introduce overall distortions or artifacts, resulting in unrealistic or unnatural results. To address this, we use GPTEval3D, which leverages GPT-4V and customizable 3D-aware prompts to offer flexible comparisons between two 3D assets based on a set of specific evaluation criteria. For more details about these metrics, please refer to Sec.~\ref{app:metrics}.

\subsection{Results}

\begin{figure*}[t]
    \centering
    \includegraphics[width=\linewidth]{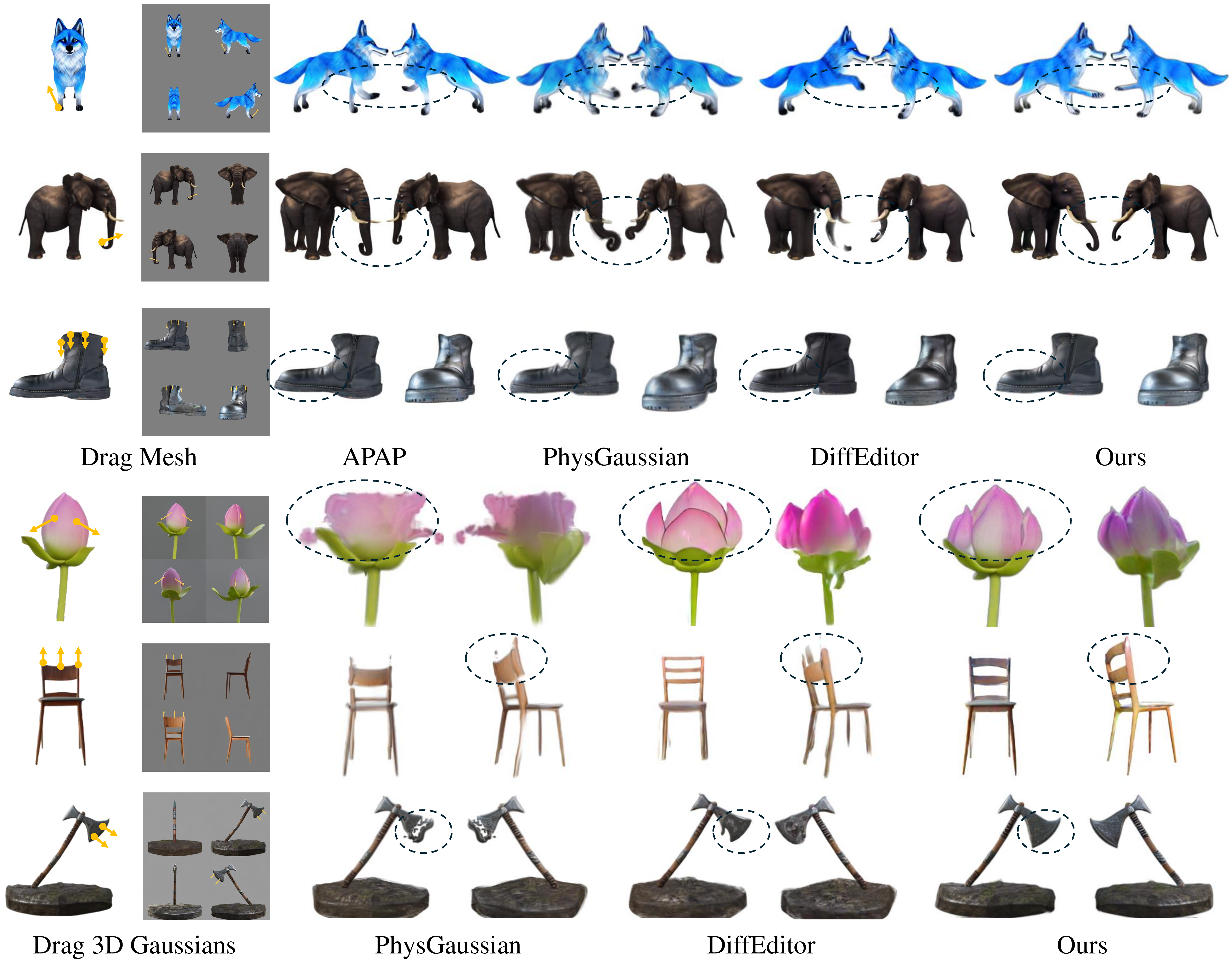}
    \caption{3D dragging results on meshes and 3D Gaussians. The first three rows show the results for the mesh, and the last three rows show the results for the 3D Gaussians. Black dashed circles indicate some detailed differences.}
    \label{fig:visual_comparison}
\end{figure*}

\textbf{Baselines}. One baseline comparison involves leveraging a 2D drag method to edit each view independently. In this setup, we use DiffEditor~\citep{mou2024diffeditor} to drag the four rendered views, followed by the same reconstruction and optimization steps as ours to produce the final 3D results. During our initial experiments, we observed that when editing much more than four views, such as 120, DiffEditor introduced significant 2D inconsistencies. Thus, for a fair comparison, we limit the process to four images as in our approach. 
We also compare our method with APAP, the state-of-the-art drag-based mesh deformation technique. Additionally, we include PhysGaussian~\citep{xie2024physgaussian}, which enables user control over Gaussian-based dynamics. For this comparison, we start with a 3D model, render four images, reconstruct a 3D Gaussian, and feed it into the PhysGaussian simulator. More detailed drag setup for PhysGaussian please refer to Sec.~\ref{app:physG}. 
%Note that we encountered issues when attempting to run Interactive3D~\citep{dong2024interactive3d}, and as a result, it could not be executed successfully. Therefore, we are unable to include it in our comparisons.
Note that as the released code of Interactive3D~\citep{dong2024interactive3d} cannot be run successfully, we are unable to include it in our comparisons. But conceptually, our approach provides a stronger multi-view diffusion prior compared to the SDS loss in Interactive3D, as we can also observe in our comparison with APAP.

\textbf{Visual Comparisons}. We first conduct a visual comparison of the proposed MVDrag3D against baselines, as demonstrated in Fig.~\ref{fig:visual_comparison}. The first three rows present results of dragging on meshes, while the last three rows show results on 3D Gaussians. For each method, we render two views to highlight the respective editing results. Take the wolf mode in the first row as an example, we aim to lift its left leg. While APAP deforms the leg, it bends rather than lifts it, resulting in a less realistic motion. In contrast, our method produces an articulation-like motion that is more natural. DiffEditor generates a successful edit in some views, but others fail, leading to inconsistent 3D results. As for PhysGaussian, it relies on predefined physical properties. Since the optimal parameters are unknown, its results exhibit some distortion. Additionally, it is unable to generate new content. For more visual results, please refer to the supplemental video demo.

\begin{table}[t]
\centering
\caption{Quantitative comparison with state-of-the-art methods on both meshes and 3D Gaussians. Left side of ``/'': Mesh. Right side: 3D Gaussians. $\gamma$ represents the patch radius, which defines the neighborhood around the 2D dragging points. APAP was not tested on 3D Gaussians. In the last column, we report a rough average running time. }
\vspace{-0.2cm}
\resizebox{1.0\textwidth}{!}{
\begin{tabular}{llllll|l}
\toprule
Method                & $\gamma=1 (\downarrow) $                          & $\gamma=3 (\downarrow) $                           & $\gamma=5 (\downarrow) $                           & $\gamma=7 (\downarrow) $                                    & $\gamma=10 (\downarrow) $          &   Time              \\
\midrule 													
APAP          & 0.2154 / -- & 0.2467 / -- & 0.2150 / -- & 0.1859 / --      & 0.1672 / -- & 6 minutes   \\
PhysGaussian  & 0.1763 / 0.2468  & 0.1887 / 0.2331    & 0.1671 / 0.2153 &  0.1448 / 0.1979      &  0.1296 / 0.1814 & 1 minutes  \\
DiffEditor    & 0.1564 / 0.1722  &  0.1452 / 0.1735   &  0.1348 / 0.1619 &  0.1299 / 0.1486     & 0.1300 / 0.1358 & 6 minutes  \\
Ours (LGM)    & 0.1153 / 0.1702  &  0.1080 / 0.1588      &  0.0989 / 0.1397    &  \textbf{0.0890} / 0.1260   & \textbf{0.0865} / 0.1130  & 3 minutes  \\
Ours + deformation   & \textbf{0.1121} / 0.1269  & \textbf{0.1044} / 0.1150   & \textbf{0.0975} / 0.1081  & 0.0908 / 0.1017  & 0.0881 / 0.0937 & 5 minutes   \\
Ours + deformation + SDS    & 0.1461 / \textbf{0.1159}   & 0.1292 / \textbf{0.1074}	    & 0.1175 / \textbf{0.1020}     & 0.1064 / \textbf{0.0960}  & 0.0994 / \textbf{0.0900}	& 8 minutes \\        
\bottomrule                    
\end{tabular}}
\vspace{-0.2cm}
\label{tab:dai}
\end{table}

\begin{table}[t]
\setlength{\tabcolsep}{1.5mm}
\caption{Evaluation results of GPTEval3D. ``Ours + deformation + SDS'' performs almost the best across all criteria on both meshes and 3D Gaussians.}
\vspace{-0.2cm}
\resizebox{1.0\textwidth}{!}{
\begin{tabular}{lcccccccccccc}
\toprule
Method                   & \multicolumn{2}{c}{\begin{tabular}[c]{@{}c@{}}Text-Asset\\ Alignment ($\uparrow$)\end{tabular}}  & \multicolumn{2}{c}{3D Plausibility ($\uparrow$)} & \multicolumn{2}{c}{\begin{tabular}[c]{@{}c@{}}Text-Geometry\\ Alignment ($\uparrow$)\end{tabular}} & \multicolumn{2}{c}{Texture Details ($\uparrow$)}  & \multicolumn{2}{c}{Geometry Details ($\uparrow$)}\hspace{-0.3cm} & \multicolumn{2}{c}{Overall ($\uparrow$)}  \\
\midrule
                     & Mesh             & 3DGS               & Mesh             & 3DGS              & Mesh             & 3DGS                  & Mesh             & 3DGS               & Mesh             & 3DGS          & Mesh             & 3DGS         \\
APAP                     & 895.53             & --                 & 906.63           & --                & 961.97              & --                   & 945.32               & --          & 905.80           & --                & 917.80           & --                \\
PhysGaussian             & 828.46             & 973.08            & 870.32           & 881.52           & 911.28              & 950.91              & 920.78               & 977.59     & 898.65           & 968.70           & 891.62           & 979.76           \\
DiffEditor               & 982.32             & 883.25            & 1054.11          & 924.96           & 1045.48              & 868.99              & 1042.24              & 894.55     & 975.34           & 885.61           & 992.50           & 897.78           \\
Ours (LGM)               & 1074.58            & 1047.74           & 1001.04          & 975.45           & \textbf{1090.78}    & 1011.64             & 1075.72              & 959.59     & 1084.85          & 1026.61          & 1041.38          & 1048.89          \\
Ours + deformation       & 1023.55            & 954.67            & 1060.81          & 947.32           & 1012.23             & 961.58              & 945.32               & 1066.18    & 1051.28          & 962.77           & 1066.18          & 982.10           \\
Ours + deformation + SDS\hspace{-0.2cm} & \textbf{1172.77}   & \textbf{1113.36}  & \textbf{1139.37} & \textbf{1103.98} & 1059.67             & \textbf{1122.44}    & \textbf{1076.25}     & \textbf{1098.33}    & \textbf{1109.46} & \textbf{1108.64} & \textbf{1136.80} & \textbf{1100.33} \\
\bottomrule
\end{tabular}}
\label{tab:GPTeval}
\end{table}

\textbf{Quantitative Comparisons}.
In addition to the visual comparisons, we conducted a quantitative evaluation to assess the effectiveness of all compared methods in terms of dragging accuracy (\textbf{DAI}) and overall editing quality (\textbf{GPTEval3D}). 
Table~\ref{tab:dai} reports different methods' DAI across varying patch radius values $\gamma$. As $\gamma$ increases from 1 to 10, our method, both with and without SDS, shows consistently lower error against other approaches like APAP, PhysGaussian, and DiffEditor. In Table~\ref{tab:GPTeval}, the GPTEval3D evaluation reveals that the ``Ours + deformation + SDS'' method performs almost the best across all criteria on both meshes and 3D Gaussians. Notably, we observed that while the SDS version of our method may not always achieve the highest DAI score, this is understandable. The SDS tends to sharpen visual details, which can lead to minor numerical decreases, but it ultimately results in more visually pleasing outputs. This is further supported by the GPTEval3D results, where the SDS version achieves the highest score in texture details.

\subsection{Abalation and Discussion}

\textbf{Abalation}. We start with the initial reconstruction from~\citep{tang2024lgm} as a baseline (Ours (LGM)) and progressively integrate our two-step optimizations: (i) Gaussian position optimization (Ours + deformation), and (ii) image-conditioned multi-view SDS (Ours + deformation + SDS). Table~\ref{tab:dai} presents a clear comparison of the impact of each stage on both mesh data and 3D Gaussians. Fig.~\ref{fig:deformation} and Fig.~\ref{fig:sds} also visually demonstrate the effectiveness of our proposed optimization strategy. %Notice that we both observe 

\begin{figure*}[t]
    \centering
    \includegraphics[width=\linewidth]{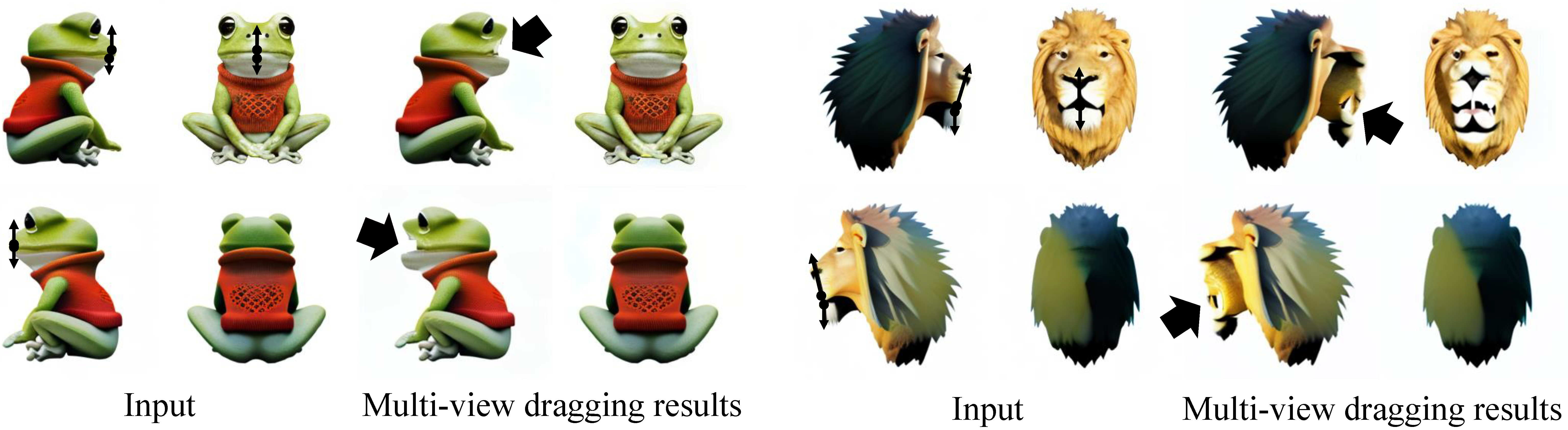}
    \vspace{-0.6cm}
    \caption{Results of dragging on image-conditioned multi-view diffusion model. We extend the dragging stage to ImageDream~\citep{wang2023imagedream}. The results are less flexible as indicated by black arrows.}
    \vspace{-0.2cm}
    \label{fig:imagecondition}
\end{figure*}

\textbf{Drag on image-conditioned diffusion model}. Considering the existence of several image-conditioned multi-view diffusion models, such as Imagedream~\citep{wang2023imagedream} and Zero123++~\citep{shi2023zero123++}, an intuitive idea is to extend the multi-view dragging stage to these models. Here, we specifically extend it to Imagedream. Fig.~\ref{fig:imagecondition} shows two cases. The conditioning image is the front view of each input. Under this setting, we observe that the results are less visually pleasing. We suspect the reason is that the image condition is too strong, thereby restricting the editing effects. In~\cite{mou2024diffeditor}, the authors introduce the use of both image and text for fine-grained image editing by tuning a new encoder, enabling a more detailed description of the desired changes. We see this as a potential direction for our work, aiming to enhance precision and flexibility in multi-view editing.

\section{Conclusion}
In this work, we introduce MVDrag3D, a novel paradigm that harnesses the power of multi-view generation-reconstruction priors for creative 3D editing. 
MVDrag3D first applies a multi-view dragging technique to ensure consistent edits across four orthogonal views. Following this, a reconstruction model generates 3D Gaussians of the edited object. To refine these initial 3D Gaussians, we introduce a deformation network that aligns the Gaussians across different views, complemented by a multi-view score function to enhance visual sharpness and consistency. Extensive experiments showcase the precision, generative capabilities, and flexibility of our method, making it a versatile solution for 3D editing across various object categories and representations. 

%Lastly, MVDrag3D is compatible with existing 3D generation or reconstruction models, allowing seamless integration—once a 3D model is generated or reconstructed, it can be edited using our approach.

\bibliography{iclr2025_conference}
\bibliographystyle{iclr2025_conference}

\appendix
\section{Appendix}
\subsection{Additional Parameters for multi-view dragging}
For multi-view image dragging, parameters such as the editing and content energy balance weights $\alpha$ and $\beta$ (see Eq.~\ref{eq:self_guidance_energy}) and the classifier-free guidance (CFG) need to be configured. We leave these as open parameters for users, as the optimal settings may vary depending on the specific edit target.

\subsection{Metric explanation}
\label{app:metrics}
\textbf{DAI}. DAI measures the effectiveness of a method in transferring semantic content to a target point. Specifically, it evaluates whether the content at the source position denoted as $\boldsymbol{p}_j$, has been successfully moved to the target location $\boldsymbol{q}_j$ in the edited 3D object.
For each 3D object, the DAI is computed over four views and considers all non-occluded dragging points as follows:
\begin{equation} \label{eq:DAI}
    {\rm DAI} =  \dfrac{1}{4} \sum_{i=1}^{4} \sum_{j=1}^{k} \dfrac{\left\Vert {\mathbf{I}_{i} \cdot \mathrm{\Omega}(\boldsymbol{p}_{i,j}^{2D},\gamma) - \mathbf{I}_{e,i} \cdot \mathrm{\Omega}(\boldsymbol{q}_{i,j}^{2D},\gamma)}\right\Vert_2^2}{(1+2\gamma)^2},
\end{equation}
where $\mathrm{\Omega}(\boldsymbol{p}_{i,j}^{2D},\gamma)$ represents a patch centered at $\boldsymbol{p}_{i,j}^{2D}$ with radius $\gamma$.
Eq.~\ref{eq:DAI} calculates the mean squared error between the patch at $\boldsymbol{p}_j^{2D}$ of $\mathbf{I}$ and the patch at $\boldsymbol{q}_j^{2D}$ of $\mathbf{I}_e$. By adjusting the radius $\gamma$, the metric can focus on different levels of context. A smaller $\gamma$ provides a precise evaluation of differences at the exact control points, while a larger $\gamma$ includes a broader region, allowing for an assessment of the surrounding context. This adaptability makes DAI a flexible tool for examining various aspects of editing quality. Given that the image resolution is $256\times256$, we set $\gamma={1,3,5,7,10}$.

\textbf{GPTEval3D}. While DAI effectively measures drag accuracy, it is not sufficient on its own because the editing process can introduce distortions or artifacts, leading to unrealistic or unnatural results. Therefore, evaluating the naturalness and fidelity of the edited images is crucial for a comprehensive quality assessment. This task is particularly challenging due to the absence of ground-truth edited 3D objects for reference.
To address this, we utilize GPTEval3D, which leverages GPT-4V with customizable 3D-aware prompts. GPTEval3D aligns well with human judgment across several dimensions, including text-to-asset alignment, 3D plausibility, texture-–geometry coherence, texture details, and geometry details. Specifically, GPTEval3D prompts GPT-4V to compare two 3D assets generated by different methods using four rendered images and normal maps. The pairwise comparisons are then used to calculate Elo ratings, which reflect each method's performance. For more details, please refer to~\citep{wu2024gpt}.

\begin{figure*}[t]
    \centering
    \includegraphics[width=\linewidth]{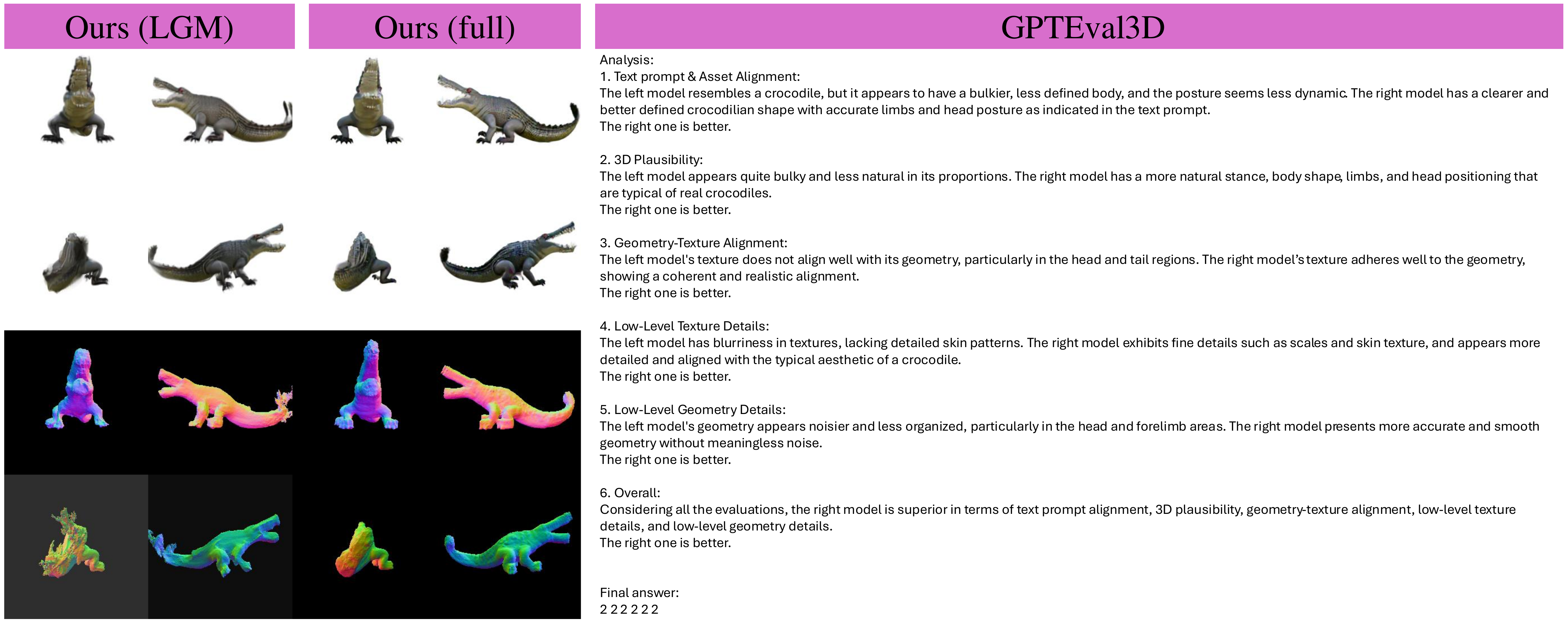}
    \caption{An analysis example of GPTEval3D on two versions of our method: Ours (LGM) and the full version, Ours + deformation + SDS. The left side of the figure shows selected four-view results from both methods, including both the appearance image and the normal map. On the right, GPT-4V's evaluation is presented, which aligns with human observations. The final line on the right confirms that the second method, Ours + deformation + SDS, outperforms the first, Ours (LGM), across all five evaluation criteria.}
    \label{fig:GPTEval3D}
\end{figure*}

Fig.~\ref{fig:GPTEval3D} presents a pairwise comparison example of GPTEval3D on two versions of our method: Ours (LGM) and the full version, Ours + deformation + SDS. The visual results on the left show that Ours (LGM) produces somewhat blurry output with noticeable noise in the geometry, particularly around the tail region. This can be attributed to the lack of optimization provided by the deformation network and SDS in this version. On the right side of the figure, GPT-4V’s judgment aligns with our observations, concluding that the second method, Ours + deformation + SDS, outperforms Ours (LGM) across all five evaluation criteria.

\subsection{Drag setup for PhysGaussian}
\label{app:physG}
In PhysGaussian~\citep{xie2024physgaussian}, we use the translation function as a proxy for the drag operation. We set the drag starting points as the center points and use the direction from the starting points to the destination points to define the initial velocity. For each dragging point pair, we assign a translation movement, and the simulation continues until either the starting point reaches the destination or the iteration count reaches the set maximum (75 by default).

\begin{wrapfigure}{r}{0.5\textwidth}
\vspace{-6mm}
    \begin{center}
       \includegraphics[width=\linewidth]{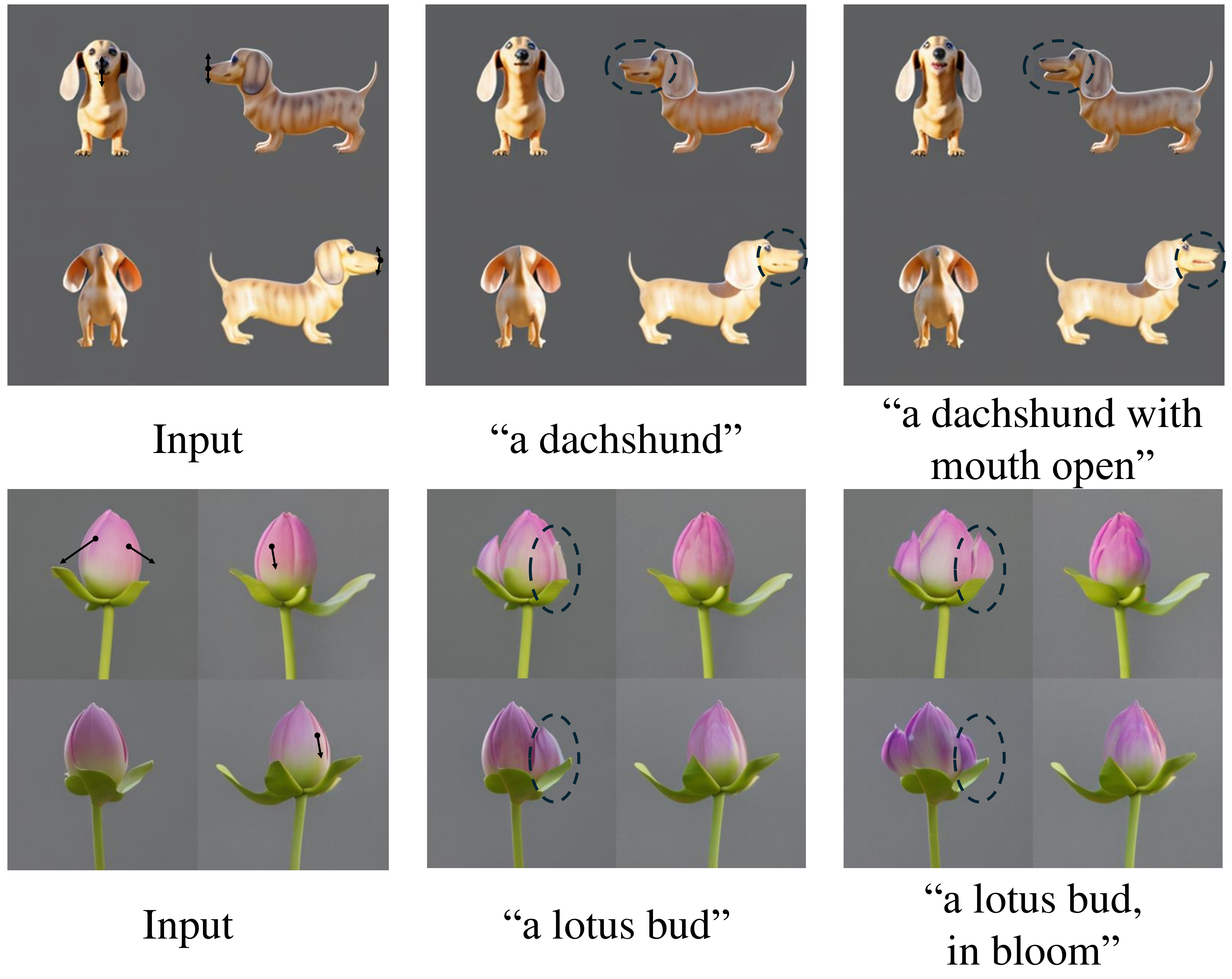}
        \caption{Effect of different text prompts. When editing images, a text prompt that better aligns with the drag intention can help query more meaningful features from the diffusion model, ultimately leading to more visually pleasing results. Black dashed circles highlight edit differences.}
        \label{fig:text_prompt}
    \end{center}
\end{wrapfigure}
\subsection{Running time statistics} 
The last column of Table~\ref{tab:dai} also summarizes the rough average running time for each method. APAP, DiffEditor, and the full version of our method are slower than PhysGaussian, Ours (LGM), and ``Ours + deformation'', mainly due to the absence of SDS optimization in their pipelines. PhysGaussian runs the fastest since it does not involve any optimization process.

\subsection{Text prompt}
Interestingly, during our early tests, we observed that text input plays a crucial cue for generative editing. As shown in Fig.~\ref{fig:text_prompt}, when dragging the dog's mouth to open, using a more specific text prompt like ``a dachshund with an open mouth'' can effectively guide the process. This proves the significance of prompt design in aligning the diffusion model’s features with the intended edits. In all our experiments, we provide a more detailed text prompt when the drag intention is clear. However, for cases where the intention is less defined, we use a more general description instead.

\begin{figure*}[t]
    \centering
    \includegraphics[width=\linewidth]{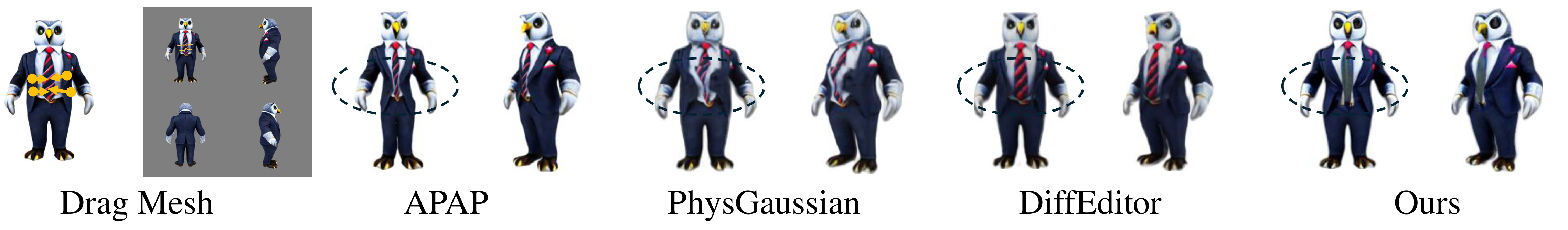}
    \caption{An example of local identity change. In this example, our goal is to drag the owl suit. Although our method successfully closes the suit, the tie part of the suit changes during the multi-view dragging process, as shown in the dashed circle region.}
    \label{fig:limitation}
\end{figure*}

\subsection{Limitations}
Despite achieving consistent results, the four-view image editing process sometimes requires significant parameter tuning, highlighting the need for a simpler, more user-friendly multi-view editing tool, akin to InstantDrag~\citep{shin2024instantdrag}. Additionally, the editing quality can occasionally alter the object's identity (the tie part of the owl suit in Fig.~\ref{fig:limitation}), how to achieve more precise local control is non-trivial. Finally, while we use multi-view images as a 3D proxy, dragging points can sometimes become occluded in all views. This limitation motivates future work on training a ``pure'' 3D generative model to enable more flexible and accurate 3D editing.

\end{document}